\documentclass{easychair}

\usepackage[utf8]{inputenc}
\usepackage{lmodern}
\usepackage{alltt} 
\usepackage{amsthm}
\usepackage{booktabs}

\usepackage{tikz}
\usetikzlibrary{shapes,arrows}
\usepackage{pgfplots}
\usepackage{xcolor}
\usepackage{url}
\usepackage{subfigure}

\usepackage{xspace}
\usepackage{amsmath}
\usepackage{listings}
\def\holfour{\textsf{HOL4}\xspace}
\def\isabelle{\textsf{Isabelle}\xspace}
\def\hollight{\textsf{HOL Light}\xspace}

\def\coq{\textsf{Coq}\xspace}

\def\eprover{\textsf{E-prover}\xspace}

\def\sml{\textsf{SML}\xspace}
\def\polyml{\textsf{Poly/ML}\xspace}
\def\holyhammer{\textsf{HOL(y)Hammer}\xspace}

\def\metis{\textsf{Metis}\xspace}

\def\tactictoe{\textsf{TacticToe}\xspace}
\newcommand{\bq}[1]{\textbackslash"{#1}\textbackslash"}
\newcommand{\ra}[1]{\renewcommand{\arraystretch}{#1}}
\theoremstyle{remark}
\newtheorem{example}{Example} 
\newtheorem{remark}{Remark} 
\usepackage{fancyvrb}
\usepackage{tikz} 
\usetikzlibrary{arrows,shapes}
\usetikzlibrary{trees,positioning,fit}
\tikzstyle{block} = [rectangle, draw,
    text width=5em, text centered, rounded corners, minimum height=4em]
\tikzstyle{line}=[draw]
\tikzstyle{cloud} = [draw, ellipse,
    node distance=3cm,
    minimum height=2em]

\title{TacticToe: Learning to reason with HOL4 Tactics}

\author{Thibault Gauthier\inst{1} \and 
Cezary Kaliszyk\inst{1} \and Josef Urban\inst{2}}

\institute{
  University of Innsbruck\\
  \email{\{thibault.gauthier,cezary.kaliszyk\}@uibk.ac.at}
\and
  Czech Technical University, Prague.\\
  \email{josef.urban@gmail.com}\\
}

\authorrunning{Gauthier, Kaliszyk and Urban}

\titlerunning{TacticToe: Learning to Reason with HOL4 Tactics}

\begin{document}
\maketitle
\begin{abstract}



Techniques combining machine learning with translation to automated
reasoning have recently become an important component of formal proof assistants. 
Such ``hammer'' techniques complement traditional proof
assistant automation as implemented by tactics and decision
procedures. In this paper we present a unified proof assistant
automation approach which attempts to automate the selection of
appropriate tactics and tactic-sequences combined with an optimized
small-scale hammering approach.
We implement the technique as a tactic-level automation for HOL4:
TacticToe. It implements a modified A*-algorithm directly in HOL4 that
explores different tactic-level proof paths, guiding their selection
by learning from a large number of previous tactic-level
proofs. Unlike the existing hammer methods, TacticToe avoids
translation to FOL, working directly on the HOL level. By combining tactic prediction and premise selection, TacticToe is able to re-prove 39\% of 7902 HOL4 theorems in 
5 seconds whereas the best single HOL(y)Hammer strategy solves 32\% in the same amount of time.

\end{abstract}

\section{Introduction}

\begin{example} Proof automatically generated by \tactictoe for the user given goal
\small
\begin{center}

\begin{BVerbatim}[commandchars=\\\{\}]
Goal: ``\(\forall\)l. FOLDL (\(\lambda\)xs x. SNOC x xs) [] l = l``
Proof:
  SNOC_INDUCT_TAC THENL
  [ REWRITE_TAC [APPEND_NIL, FOLDL],
    ASM_REWRITE_TAC [APPEND_SNOC, FOLDL_SNOC]
      THEN CONV_TAC (DEPTH_CONV BETA_CONV)
      THEN ASM_REWRITE_TAC [APPEND_SNOC, FOLDL_SNOC] ]
\end{BVerbatim}

\end{center}

\end{example}

  Many of the state-of-the-art interactive theorem provers (ITPs) such as 
  \holfour~\cite{hol4}, \hollight~\cite{Harrison09hollight}, \isabelle~\cite{isabelle} 
  and \coq~\cite{coq-book}
  provide high-level parameterizable tactics for constructing the
  proofs.  Such tactics typically analyze the current goal state and
  assumptions, apply nontrivial proof transformations, which get
  expanded into possibly many basic kernel-level inferences or significant parts of the proof term.
  In this
  work we develop a tactic-level automation procedure for the \holfour ITP
  which guides selection of the tactics by learning from previous
  proofs.  Instead of relying on translation to first-order automated
  theorem provers (ATPs) as done by the hammer systems~\cite{hammers4qed,tgck-cpp15}, the technique
  directly searches for sequences of tactic applications that lead to
  the ITP proof, thus avoiding the translation and
  proof-reconstruction phases needed by the hammers.

  To do this, we \emph{extract and record} tactic invocations from the ITP proofs (Section~\ref{sec:recording}) and
  \emph{build efficient machine learning classifiers} based on such training
  examples (Section~\ref{sec:selecting}).  The learned data serves as a guidance for our \emph{modified
  A*-algorithm} that explores the different proof paths (Section~\ref{sec:proofsearch}). The result, if
  successful, is a certified human-level proof composed of \holfour
  tactics.  The system is evaluated on a large set of theorems originating
  from \holfour (Section~\ref{sec:setting}), and we show that the performance of the single best \tactictoe strategy exceeds the performance of a hammer system used
  with a single strategy and a single efficient external prover.

\paragraph{Related Work}

There are several essential components of our work that are comparable to previous approaches: tactic-level proof recording, tactic 
selection through machine learning techniques and automatic tactic-based proof search. Our work is also related to previous approaches that use machine learning to select premises for the ATP systems and guide ATP proof search internally.

For \hollight, the Tactician tool~\cite{DBLP:conf/sefm/Adams15} 
can transform a packed tactical proof into a series of interactive tactic calls. Its principal application 
was so far refactoring the library and teaching common proof techniques to new ITP users. In our work, the splitting of a proof into a sequence of tactics is essential for the
tactic recording procedure, used to train our tactic prediction module.

The system \textsf{ML4PG}~\cite{DBLP:journals/corr/abs-1212-3618, DBLP:journals/mics/HerasK14} 
groups related proofs thanks to its clustering 
algorithms. It allows \coq users to inspire themselves from similar proofs and notice 
duplicated proofs. Our predictions comes from a much more detailed description of the open goal.
However, we simply create a single label for each tactic call whereas each of its
arguments is treated independently in \textsf{ML4PG}. 
Our choice is motivated by the k-NN algorithm already used in
\holyhammer for the selection of theorems.

\textsf{SEPIA}~\cite{DBLP:conf/cade/GransdenWR15} is a powerful system able to generate
proof scripts from previous \coq proof examples.
Its strength lies in its ability to produce likely sequences 
of tactics for solving domain specific goals. It operates by creating a model for common sequences of tactics for a specific library.
This means that in order to propose the following tactic, only the previously called tactics
are considered.
Our algorithm, on the other hand, relies mainly on the characteristics of the current goal 
to decide
which tactics to apply next. In this way, our learning mechanism has to rediscover why each 
tactic was applied for the current subgoals. It may lack some useful bias for common sequences 
of tactics, but is more reactive to subtle changes. Indeed, it can be trained on a large library and only tactics relevant to the current subgoal will be selected. 
Concerning the proof search, \textsf{SEPIA}'s 
breadth-first search is replaced by an A*-algorithm which allows for heuristic guidance in 
the exploration of the search tree.
Finally, \textsf{SEPIA} was evaluated on three chosen parts (totaling 2382 theorems) of the 
\coq library demonstrating that it globally outperforms individual \coq tactics. In contrast, we demonstrate the competitiveness of our system against the successful general-purpose hammers on the \holfour standard library (7902 theorems).

Machine learning has also been used to advise the best library lemmas for new ITP goals.
This can be done either in an interactive way, when the user completes the proof based on the recommended lemmas, as in the \textsc{Mizar Proof Advisor}~\cite{Urb04-MPTP0}, or attempted fully automatically, where such lemma selection is handed over to the ATP component of a \emph{hammer} system~\cite{hammers4qed,tgck-cpp15,holyhammer,BlanchetteGKKU16,mizAR40}.

Internal learning-based selection of tactical steps inside an ITP is analogous to internal learning-based selection of clausal steps inside ATPs such as \textsc{MaLeCoP}~\cite{malecop} and \textsc{FEMaLeCoP}~\cite{femalecop}. These systems
use the naive Bayes classifier to  select clauses for the extension steps in
tableaux proof search based on many previous proofs. Satallax~\cite{Brown2012a} can guide its
search internally~\cite{mllax} using a command classifier, which can estimate the priority of the 11 kinds of
commands in the priority queue based on positive and negative examples.


\section{Recording Tactic Calls}\label{sec:recording}

Existing proof recording for \holfour~\cite{Wong95recordingand,DBLP:conf/itp/KumarH12}
relies on manual modification of all primitive inference rules in the kernel.
Adapting this approach to record tactics would require the manual 
modification of the 750 declared \holfour tactics.
Instead, we developed an
automatic transformation on the actual proofs. Our process
singles out tactic invocations and introduces calls to general purpose recording in the proofs.
The main benefit of our approach is an easy access to the string representation of the tactic and its 
arguments which is essential to automatically construct a human-level proof 
script. As in the LCF-style theorem prover users may introduce new tactics or arguments
 with the \texttt{let} construction inline, special care needs to be taken so 
 that the
tactics can be called in any other context. The precision of the recorded information
will influence the quality of the selected tactics in later searches.
The actual implementation details of the
recording are explained in the Appendix. 

%

\section{Predicting Tactics}\label{sec:selecting}

The learning-based selection of relevant lemmas significantly improves the automation for hammers~\cite{BlanchetteGKKU16}.
Therefore we propose to adapt one of the strongest hammer lemma selection methods to predict tactics 
in our setting: the modified distance-weighted \emph{$k$ nearest-neighbour} (k-NN) classifier~\cite{ckju-pxtp13,DudaniS76}.
Premise selection usually only prunes the initial set of formulas given to the ATPs, which then try to solve the pruned problems on their own. 
Here we will use the prediction of relevant tactics to actively guide 
the proof search algorithm (described in Section~\ref{sec:proofsearch}).

Given a goal $g$, the classifier selects a set of previously solved goals similar to $g$, and considers
the tactics that were used to solve these goals as relevant for $g$.
As the predictor bases its relevance estimation on frequent similar goals, it is crucial to estimate the distance between
the goals in a mathematically relevant way. We will next discuss the extraction of the features from the goals and the
actual prediction. Both have been integrated in the \sml proof search.

\subsection{Features}\label{sec:features}

We start by extracting the syntactic features that have been successfully used in premise selection from the goal:
\begin{itemize}
\item names of constants, including the logical operators,
\item type constructors present in the types of constants and variables,
\item first-order subterms (fully applied) with all variables replaced by a single place holder $V$.
\end{itemize}

We additionally extract the following features:

\begin{itemize}
\item names of the variables,
\item the top-level logical structure with atoms substituted by a single place 
holder $A$ and all its substructures.
\end{itemize}

We found that the names of variables present in the goal are particularly important for tactics such as 
case splitting on a variable 
(\texttt{Cases\_on var}) or instantiation of a variable 
(\texttt{SPEC\_TAC var term}). Determining the presence of top-level logical operators (i.e 
implication) is essential to assess if a "logical" tactic should be applied. For example, 
the presence of an implication may lead to the application of the tactic \texttt{DISCH\_TAC} that moves the precondition to the assumptions. Top-level logical structure gives a more detailed view of the relationship between those logical components. Finally, we also experiment with some general features because they are natural in higher-order logic:
\begin{itemize}
\item (higher-order) subterms with all variables unified, including partial function applications.
\end{itemize}

\subsection{Scoring}\label{sec:predictions}

In all proofs, we record each tactic invocation and link (associate) the currently open goal with the tactic's name in our database.
Given a new open goal $g$, the \emph{score of a tactic $T$ wrt. $g$} is defined to be the score (similarity) of $T$'s 
associated goal which is most similar to $g$.
The idea is that tactics with high scores will be more likely to solve the open goal $g$, 
since they were able to solve similar goals before.

We estimate the similarity (or co-distance) between an open goal $g_o$ and a previously recorded goal $g_p$ using their respective feature sets $f_o$ and $f_p$.
The co-distance $tactic\_score_1$ computed by the k-NN algorithm is analogous to the
one used in the premise selection task~\cite{ckju-pxtp13}. The main idea 
is to find the features shared by the two goals and estimate the rarity of 
those features calculated via the TF-IDF~\cite{Jones72astatistical} heuristics.
In a second co-distance $tactic\_score_2$, we additionally take into account the
total number of 
features to reduce the seemingly unfair advantage of big feature sets in the 
first scoring function.
\[tactic\_score_1 (f_o, f_p) = {\sum\nolimits_{\,f \in f_o \cap f_p}{\text{tfidf}(f)^{\tau_1}}}\]

\[tactic\_score_2 (f_o,f_p) = \frac{tactic\_score_1 (f_o,f_p)}{(1+ ln\ (1 + card\ f_o))} \]

Moreover, we would like to compare the distance of a recorded goal to 
different goals opened at different moment of the search. That is why we 
normalize the scores by dividing them by the similarity of the open goal with 
itself. As a result, every score will lie in the interval $[0, 1]$ where 1 is 
the highest degree of similarity (i.e. the shortest distance). We respectively 
refer to those normalized scores later as $tactic\_norm\_score_1$ and $tactic\_norm\_score_2$.

\subsection{Preselection}\label{sec:pretactic} 

Since the efficiency of the predictions will be crucial during the proof 
search, we preselect 500 tactics before the search.
A sensible approach here is to preselect a tactic based on the distance 
between the statement of the conjecture to be proven and the statement(s) for which the tactic is 
part of the proof. During the proof, when an open goal is created, only the scores
of the 500 preselected tactics will be recalculated and the tactics will be 
reordered according to these scores.

\subsection{Orthogonalization}
Different tactics may transform a single goal in the same way. Exploring such equivalent paths
is undesirable, as it leads to inefficiency in automated proof search.
To solve this problem, we do not directly assign a goal to the associated 
tactic, but organize a competition on the closest feature vectors
(tactic string together with the features of an associated goal). The winner 
is the tactic appearing in the most feature vectors provided that it has the same effect as the original tactic. We associate this tactic with the features of the targeted goal instead of the original in our feature database. As a result, already 
successful tactics are preferred, and new tactics are considered only if they 
provide a different contribution.

\subsection{Self-learning}
If the search algorithm finds a proof, we record both the human and computer-generated proof in the
feature database. Since recording and re-proving are intertwined, the 
additional data is available for the next proof search.
The hope is that it will be easier for \tactictoe to learn from its own 
discovered proofs than from the human proof scripts~\cite{DBLP:conf/cade/Urban07}.

\section{Proof Search Algorithm}\label{sec:proofsearch}

Despite the best efforts of the prediction algorithms, the selected tactic may not 
solve the current goal, proceed in the wrong direction
or even loop. For that reason, the prediction needs to be accompanied by a  
proof search mechanism that allows for backtracking and 
can choose which proof tree to extend next and in 
which direction.

Our search algorithm takes inspiration from the A*-algorithm~\cite{HartNR68} which uses 
a cost function and heuristics to estimate the total distance to the destination and
choose the shortest route. The first necessary adaptation of the algorithm stems
from the fact that a proof is in general not a
path but a tree. This means that our search space has two branching factors: the 
choice of a tactic, and the number of goals produced by tactics.
The proof is not finished when the current tactic solves its goal because it often leaves
new pending open goals along the current path.

\paragraph{Algorithm Description}

%
%

\begin{figure}{}
\centering

  \begin{subfigure}{}
  \begin{tikzpicture}[auto]
    \node [cloud] (C) {conjecture};
    \coordinate [below of=C] (M);
    \node [cloud,left of=M,draw=none] (L) {};
    \node [cloud,right of=M,draw=none] (R) {};
    \draw[-to,black,dotted,thick] (C) to [out=180,in=90] node {$tactic_1$} (L);
    \draw[-to,black,dotted,thick] (C) to [out=0,in=90] node [below,xshift=-13] {$tactic_2$} (R);
  \end{tikzpicture}
  \end{subfigure} 
  
  \begin{subfigure}{}
  \begin{tikzpicture}[auto]
    \node [cloud] (C) {conjecture};
    \coordinate [below of=C] (M);
    \node [cloud,left of=M,draw=none] (L) {\phantom{$goal_1$,$goal_2$}};
    \node [cloud,right of=M] (R) {$goal_1$,$goal_2$};
    \draw[-to,black,dotted,thick] (C) to [out=180,in=90] node {$tactic_1$} (L);
    \draw[-to,black,thick] (C) to [out=0,in=90] node [below,xshift=-13] {$tactic_2$} (R);
    \coordinate [below of=R] (M2);
    \node [cloud,left of=M2,draw=none] (L2) {\phantom{$goal_3$}};   
    \node [cloud,right of=M2,draw=none] (R2) {\phantom{$goal_3$}};
    \draw[-to,black,dotted,thick] (R) to [out=180,in=90] node {$tactic_1$} (L2);
    \draw[-to,black,dotted,thick] (R) to [out=0,in=90] node [below,xshift=-13] {$tactic_2$} (R2);
  
  \end{tikzpicture}
  \end{subfigure} 
  
  \begin{subfigure}{}
  \begin{tikzpicture}[auto]
    \node [cloud] (C) {conjecture};
    \coordinate [below of=C] (M);
    \node [cloud,left of=M,draw=none] (L) {\phantom{$goal_1$,$goal_2$}};
    \node [cloud,right of=M] (R) {$goal_1$,$goal_2$};
    \draw[-to,black,dotted,thick] (C) to [out=180,in=90] node {$tactic_1$} (L);
    \draw[-to,black,thick] (C) to [out=0,in=90] node [below,xshift=-13] {$tactic_2$} (R);
    \coordinate [below of=R] (M2);
    \node [cloud,left of=M2] (L2) {$goal_3$};   
    \node [cloud,right of=M2,draw=none] (R2) {$goal_1$ solved};
    \draw[-to,black,thick] (R) to [out=180,in=90] node {$tactic_1$} (L2);
    \draw[-to,black,thick] (R) to [out=0,in=90] node [below,xshift=-13] {$tactic_2$} (R2);
    
    \coordinate [below of=L2] (M3);    
    \node [cloud,left of=M3,draw=none] (L3) {};
    \node [cloud,right of=M3,draw=none] (R3) {};
    \draw[-to,black,dotted,thick] (L2) to [out=180,in=90] node {$tactic_1$} (L3);
    \draw[-to,black,dotted,thick] (L2) to [out=0,in=90] node [below,xshift=-13] {$tactic_2$} (R3); 
  \end{tikzpicture}
    \end{subfigure} 
  
  \begin{subfigure}{}
  \begin{tikzpicture}[auto]
    \node [cloud] (C) {conjecture};
    \coordinate [below of=C] (M);
    \node [cloud,left of=M,draw=none] (L) {\phantom{$goal_1$,$goal_2$}};
    \node [cloud,right of=M] (R) {$goal_2$};
    \draw[-to,black,dotted,thick] (C) to [out=180,in=90] node {$tactic_1$} (L);
    \draw[-to,black,thick] (C) to [out=0,in=90] node [below,xshift=-13] {$tactic_2$} (R);
    \coordinate [below of=R] (M2);
    \node [cloud,left of=M2,draw=none] (L2) {};   
    \node [cloud,right of=M2,draw=none] (R2) {};
    \draw[-to,black,dotted,thick] (R) to [out=180,in=90] node {$tactic_1$} (L2);
    \draw[-to,black,dotted,thick] (R) to [out=0,in=90] node [below,xshift=-13] {$tactic_2$} (R2);
  \end{tikzpicture}
  \end{subfigure} 
  
  \caption{4 successive snapshots of a proof attempt showing the essential steps of the algorithm: node creation, node extension and node deletion.}
  \label{fig:proof}
  \end{figure}
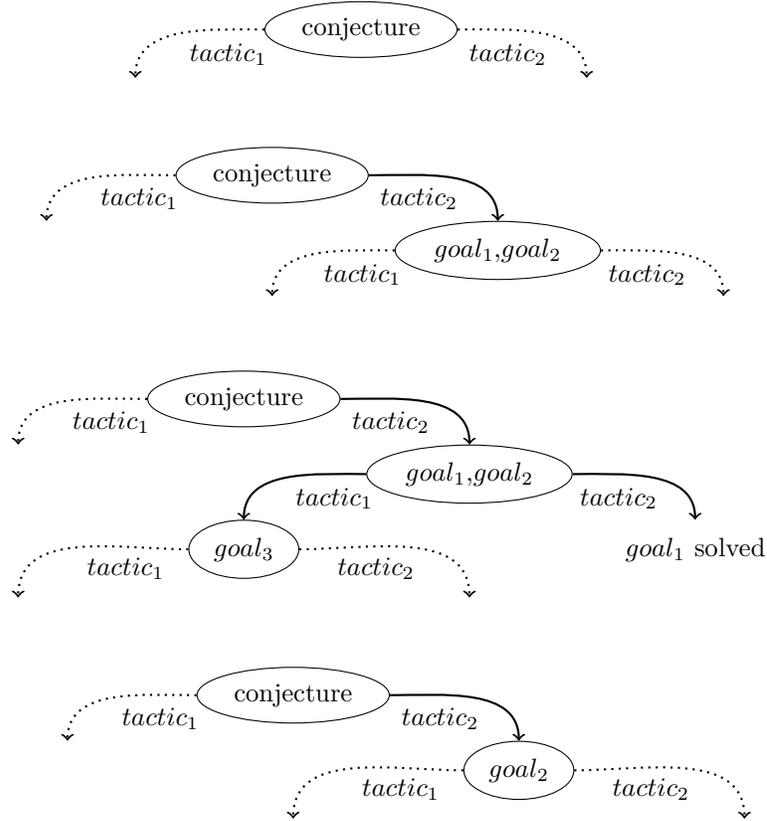

In the following, we assume that we already know the distance function (it will be defined in \ref{distancefun}) and describe how the A*-algorithm is transformed into a proof search algorithm.
In order to help visualizing the proof steps, references to a proof search example 
depicted in Figure~\ref{fig:proof} will be made throughout the description of the 
algorithm. To minimize the width of the trees in our example the branching factor is limited to two tactics $tactic_1$ and $tactic_2$ but a typical search relies on 500 preselected tactics.

Our search algorithm starts by creating a root node containing the conjecture as an open goal. A list of 500 potential tactics is attached to this node. A score for each of those 
tactics is given by the tactic selection algorithm. The tactic with the best score 
($tactic_2$ in our example) is applied to the conjecture. If no error occurs, it produces a 
new node containing a list of goals to be solved. The first of these goals ($goal_1$) is the 
open goal for the node, other goals ($goal_2$) are pending goals waiting for the first goal to be proved. 
From now, we have more than one node that can be extended, and the selection process has two steps:
First, we select the best unused tactic for each open goal ($tactic_1$ for $goal_1$, $tactic_1$ for the $conjecture$). Next, we chose the node ($goal_1$) with the highest co-distance (see next paragraph) which is supposed to be the closest to finish the proof.
The algorithm goes on creating new nodes with new open goals ($goal_3$) until a tactic ($tactic_2$) proves a goal ($goal_1$). This is the case when a tactic returns an empty list of goals or if all the goals directly produced by the tactic have already been proven.
At this point, all branches originating from the node of the solved goal are deleted and the tactic that led to the proof is saved for a later reconstruction (see Section~\ref{sec:reconstruction}). 

The whole process can stop in three different ways. 
The conjecture is proven if all goals created 
by a tactic applied to the conjecture are closed. 
The search saturates if no predicted tactics are applicable to any open goals. 
The process times out if it runs longer than the fixed time 
limit (5 seconds in our experiments).

\paragraph{Optimizations} A number of constraints are used to 
speed up the  proof search algorithm.
We forbid the creation of nodes that contain a parent goal in order to avoid loops.
We minimize parallel search by imposing that two sibling nodes must not 
contain the same set of goals. We cache the tactic applications and the 
predictions so that they can be replayed quickly if the same open goals reappears
anywhere in the search. Tactics are restricted to a very small time limit, the 
default being 0.02 seconds in our experiment. Indeed, a tactic may loop or take 
a large amount of time, which would hinder the whole search process.
Finally, we reorder the goals in each node so that the hardest goals according to the selection
heuristic are considered first. 


\subsection{Heuristics for Node Extension}\label{distancefun}

It is crucial to define a good distance function for the (modified) A*-algorithm.
This distance (or co-distance) should estimate for the 
edges of each node, how close it is to complete the proof. For the heuristic part, we rely on the score of the best tactic not yet applied to the node's first goal.
Effectively, the prediction scores evaluate a co-distance to a provable goal, with which we approximate the co-distance to a theorem.
A more precise distance estimation could be obtained
by recording the provable subgoals that have already been tried~\cite{ckju-jsc15},
however this is too costly in our setting.
We design the cost function, which represents the length of the 
path already taken as a coefficient applied to the heuristics.
By changing the parameters, we create and experiment with 5 possible co-distance functions:

\begin{align*}
codist_1 &= tactic\_norm\_score_1\\
codist_2 &= tactic\_norm\_score_2\\
codist_3(k_1) &= k_1 ^ d * tactic\_norm\_score_1\\
codist_4(k_1,k_2) &= k_1 ^ d * k_2 ^ w * tactic\_norm\_score_1\\
codist_5(k_1,k_2) &= k_1 ^ d * k_2 ^ w
\end{align*}
where $d$ is the depth of the considered node, $w$ is the number of tactics previously applied to the same goal and $k_1$, $k_2$ are coefficients in $]0, 1[$. 

\begin{remark}
If $k_1 = k_2$, the fifth co-distance has the same effect as the distance $d + w$.
\end{remark}

\paragraph{Admissibility of the Heuristic and Completeness of the Algorithm}

An important property of the A*-algorithm is the admissibility of its heuristic. A heuristic is admissible if it does not overestimate the distance to the goal. 
The fifth co-distance has no heuristic, so it is admissible.
As a consequence, proof searches based on this co-distance will find optimal solutions relative to its cost function. For the third and fourth co-distances, we can only guarantee a weak form of completeness. If there exists a proof involving the 500 preselected 
tactics, the algorithm will find one in a finite amount of time.
It is sufficient to prove that eventually the search will find all proofs at depth $\leq  k$. 
Indeed, there exists a natural number $n$, such that proofs of depth greater than $n$ have a 
cost coefficient smaller than the smallest co-distance at depth $\leq k$.
Searches based on the first two co-distances are only guided by their heuristic and therefore 
incomplete. This allows them to explore the suggested branches much deeper.

\begin{remark}
The completeness result holds only if the co-distance is positive, which happens when top-level logical structures are considered.
\end{remark}

In the future, we consider implementing the UCT-method~\cite{montecarlo} commonly used as a 
selection strategy in Monte-Carlo tree search. This method would most likely find a better balance between completeness and exploration.

\subsection{Reconstruction}\label{sec:reconstruction}
When a proof search succeeds (there are no more pending goals at the root)
we need to reconstruct a \holfour human-style proof.
The saved nodes consist of a set of trees where each edge is a tactic and
the proof tree is the one starting at the root.
In order to obtain a single \holfour proof, we need to combine the tactics
gathered in the trees using tacticals.
By the design of the search, a single tactic combinator, \texttt{THENL}, is sufficient. It combines a tactic with a list of subsequent ones, in such a way that after the parent tactic is called, for each created goal a respective tactic from the list is called.
The proof tree is transformed into a final single proof script
 by the following recursive function $P$ taking a
tree node $t$ and returning a string:
\begin{equation*}
P(t) =
\begin{cases}
P(c) & \text{if $t$ is a root},\\
tac & \text{if $t$ is a leaf},\\
tac\ \texttt{THENL}\ [P(c_0),\ldots,P(c_n)] & \text{otherwise.}
\end{cases}
\end{equation*}
where $tac$ is the tactic that produced the node, $c$ is the
only successful child of the root and $c_0, \ldots, c_n$ are the 
children of the node produced by the successful tactic.

The readability of the created proof scripts is improved, by replacing
replacing  \texttt{THENL} by \texttt{THEN} when the list has length 1.
Further post-processing such as
removing unnecessary tactics and theorems has yet to be developed but would 
improve the user experience greatly~\cite{DBLP:conf/sefm/Adams15}.

\subsection{Small ``hammer'' Approach}

General-purpose proof automation mechanisms which combine proof translation to
ATPs with machine learning (``hammers'') have become quite successful in
enhancing the automation level in proof assistants~\cite{hammers4qed}.
As external automated reasoning techniques often outperform the combined power
of the tactics, we would like to combine the \tactictoe search with \holyhammer
for \holfour~\cite{tgck-cpp15}. Moreover our approach can only use previously called
tactics, so if a theorem is essential for the current proof but has never been
used as an argument of a tactic, the current approach would fail.

Unfortunately external calls to \holyhammer at the proof search nodes are too
computationally expensive. We therefore create a ``small hammer'' comprised of a
faster premise selection algorithm combined with a short call to the internal prover
\metis~\cite{metis}. First, before the proof search, we preselect 500 theorems for the whole
proof search tree using the usual premise selection algorithm with the dependencies.
At each node a simpler selection process will select a small subset of the 500 to
be given to \metis using a fast similarity heuristic (8 or 16 in our experiment).
The preselection relies on the  theorem dependencies, which
usually benefits hammers, however for the final selection 
we only compute the syntactic feature distance works better.

During the proof search, when a new goal is created or a fresh pending goal is considered, the ``small hammer'' is always called first. Its call is associated with a tactic string for a flawless integration in the final proof script.

\section{Experimental Evaluation}\label{sec:setting}

The results of all experiments are available at:\\[1mm]
  \centerline{\url{http://cl-informatik.uibk.ac.at/users/tgauthier/tactictoe/}}
 

\subsection{Methodology and Fairness}

The evaluation imitates the construction of the library: For each theorem only the previous
human proofs are known. These are used as the learning base for the predictions.
To achieve this scenario we re-prove all theorems during a modified build of \holfour.
As theorems are proved, they are recorded and included in the training examples.
For each theorem we first attempt to run the \tactictoe search with a time limit of 5 seconds,
before processing the original proof script.
In this way, the fairness of the
experiments is guaranteed by construction. 
Only previously declared \sml 
variables (essentially tactics, theorems and simpsets) are accessible. 
And for each theorem to be re-proven \tactictoe is only trained on previous proofs.

Although the training process in each strategy on its own is fair, the selection of the 
best strategy in Section~\ref{sec:sel_param} should also be considered as a learning 
process. To ensure the global fairness, the final experiments in Section~\ref{sec:exp_full} 
runs the best strategy on the full dataset which is about 10 times larger. The
performance is minimally better on this validation set.

\subsection{Choice of the Parameters}\label{sec:sel_param}

In order to efficiently determine the contribution of each parameter,
we design a series of small-scale experiments where each evaluated strategy is run on 
every tenth goal in each theory.
A smaller dataset (training set) of 860 theorems allows testing the combinations of various
 parameters. To compare them, we propose three successive experiments that attempt to 
optimize their respective parameters. To facilitate this process further, every strategy will 
differ from a default one by a single parameter. The results will show in addition to the 
success rate, the number of goals solved by a strategy not solved by another 
strategy $X$. 
This number is called $U(X)$.

The first experiment concerns the choice of the right kind of features and feature scoring 
mechanism. The results are presented in Table~\ref{tab:feature_param}.
We observe that the higher-order features and the feature of the top logical structure 
increase minimally the number of problems solved. It is worth noting that using only first-order
features leads to 18 proofs not found by relying on additional higher-order features. 
The attempted length penalty on the total number of features is actually slightly harmful.

\begin{table}[t]
\centering\ra{1.3}
\small
\begin{tabular}{llll}
\toprule
 ID & Learning parameter & Solved & $U(D_1)$ \\
\midrule
 $D_0$ & $codist_1$ (length penalty) & 172 (20.0\%) & 5 \\
 $D_1$ & $codist_0$ (default) & 179 (20.8\%) & 0 \\
 $D_2$ & no top features & 175 (20.3\%) & 18 \\
 $D_3$ & no higher-order features & 178 (20.7\%) & 8 \\
\bottomrule
\end{tabular}
\caption{\label{tab:feature_param}Success rate of strategies with different learning parameters on the training set.}
\end{table}

In the next experiment shown in Table~\ref{tab:search_param}, we focus our attention on the search parameters. To ease comparison, we reuse the strategy $D_1$
from Table~\ref{tab:feature_param} as the default strategy.
We first try to change the tactic timeout, as certain tactics may require more time 
to complete. It seems that the initial choice of 0.02 seconds per tactic inspired by 
hammer experiments~\cite{tgckckmn-paar14} involving \metis is a good compromise. 
Indeed, increasing the timeout leaves less time for other tactics to be tried, whereas
decreasing it too much may prevent a tactic from succeeding. Until now, we trusted 
the distance heuristics completely not only to order the tactics but also to 
choose the 
next extension step in our search. We will add coefficients that reduce the scores of 
nodes deep in the search. From $D_6$ to $D_8$, we steadily increase the strength of 
the coefficients, giving the cost function of the A*-algorithm more and more influence 
on the search. The success rate increases accordingly, which means that using the 
current heuristics is a poor selection method for extending nodes. A possible solution 
may be to try to learn node selection independently from tactic selection. So it is 
not surprising that the strategy $D_9$ only relying on the cost function performs the 
best. As a minor consolation, the last column shows that the heuristic-based proof 
$D_1$ can prove 10 theorems that $D_9$ cannot prove. Nevertheless, we believe that the 
possibility of using a heuristic as a guide for the proof search is nice asset of \tactictoe.

The third experiment, presented in Table~\ref{tab:lh_param},
evaluates the effect of integrating the ``small hammer'' in the \tactictoe search.
At a first glance, the increased success rate is significant for all tested parameters. 
Further analysis reveals that increasing the number of premises from 8 to 16 with a 
timeout of 0.02 seconds is detrimental. The $D_{19}$ experiment  demonstrates that 0.1 
seconds is a better time limit for reasoning with 16 premises. And the $D_{18}$ experiment reveals the disadvantage of unnecessarily increasing the timeout of \metis. This reduces the time available for the rest of the proof search, which makes the success rate drop.

\begin{table}[t]
\centering\ra{1.3}
\small
\begin{tabular}{llll}
\toprule
 ID & Searching parameter & Solved & $U(D_{9})$ \\
\midrule
 $D_1$ & $codist_0$ (default)      & 179 (20.8\%) & 10 \\
 $D_4$ & tactic timeout 0.004 sec & 175 (20.3\%) & 9 \\
 $D_5$ & tactic timeout 0.1 sec   & 178 (20.7\%) & 10 \\
 $D_6$ & $codist_3(0.8)$     & 192 (22.3\%) & 10 \\
 $D_7$ & $codist_4(0.8,0.8)$ & 199 (23.1\%) & 7 \\
 $D_8$ & $codist_4(0.4,0.4)$ & 205 (23.8\%) & 3 \\
 $D_9$ & $codist_5(0.8,0.8)$ & 211 (24.5\%) & 0 \\
\bottomrule
\end{tabular}
 \caption{\label{tab:search_param}Success rate of strategies with different search parameters on the training set.}
\end{table}

\begin{table*}[t]
\centering\ra{1.3}
\small
\begin{tabular}{llll}
\toprule
 ID & ``small hammer'' parameter & Solved &  $U(D_{19})$ \\
\midrule
 $D_9$ & $codist_5(0.8,0.8)$ (default: no small hammer) & 211 (24.5\%) & 19 \\
 $D_{16}$ & 8 premises + timeout 0.02 sec & 281 (32.7\%) & 21 \\ 
 $D_{17}$ & 16 premises + timeout 0.02 sec & 270 (31.4\%) & 18 \\
 $D_{18}$ & 8 premises + timeout 0.1 sec & 280 (32.6\%) & 11 \\
 $D_{19}$ & 16 premises + timeout 0.1 sec & 289 (33.6\%) & 0 \\
\bottomrule
\end{tabular}
 \caption{\label{tab:lh_param}Success rate of strategies with different parameters of
 ``small hammer'' on the training set.}
\end{table*}

The best strategy which does not rely on the ``small hammer'' approach $D_9$ will be called \tactictoe(NH) (for no ``small hammer'') in the remaining part of the paper, and the best strategy relying on the approach, $D_{19}$, will be referred to as \tactictoe(SH) (``small hammer'').

\subsection{Full-scale Experiments}\label{sec:exp_full}

We evaluate the two best \tactictoe strategies on a bigger data set. Essentially, we try 
to re-prove every theorem for which a tactic proof script was provided. The 
majority of theorems in the \holfour standard library (7954 out of 10229) have been proved this 
way. The other theorems were created by forward rules and almost all of those proofs are bookkeeping operations such as instantiating a theorem or splitting a conjunction.

In addition we evaluate the two proposed more advanced strategies: self-learning and 
orthogonalization.

We will also compare the performance of \tactictoe with the \holyhammer
system \holfour~\cite{tgck-cpp15}, which has so far provided the most successful
general purpose proof automation.
Although \holyhammer has already been thoroughly evaluated, we reevaluate its best single 
strategy to match the conditions of the \tactictoe experiments. Therefore, this experiment 
is run on the current version of \holfour with a time limit of 5 seconds.
The current best strategy for \holyhammer in \holfour is using \eprover \cite{eprover,Schulz:LPAR-2013}
with the $new\_mzt\_small$ strategy discovered by \texttt{BliStr}~\cite{blistr}.
To provide a baseline the less powerful $auto$ strategy for \eprover was also tested.


The evaluation of \tactictoe is performed as part of the \holfour build 
process, whereas \holyhammer is evaluated after the complete build because of 
its export process. The consequence is that overwritten theorems are not 
accessible to \holyhammer. Conversely, each theorem which was proved directly 
through forward rules was 
not considered by \tactictoe. To estimate the relative strength of the two provers in a 
fair manner we decided to compute all subsequent statistics on the common part of the dataset. This common part consists of 7902 theorems from 134 theories.


\begin{table*}[t]
\centering\ra{1.3}
\small
\begin{tabular}{llll}
\toprule
 ID & Parameter & Solved & $U(\tactictoe(SH))$ \\ \midrule
 \tactictoe(NH)  & default                & 2349 (29.73\%)  & 173 \\
 \tactictoe(SH) & ``small hammer''       & 3115 (39.42\%)  & $U(blistr)$ : 1335 \\ 
 \tactictoe($E_2$)    & self-learn        & 2343 (29.66\%)  & 187\\
 \tactictoe($E_3$)   & self-learn + ortho & 2411 (30.51\%)  & 227\\
 \holyhammer(auto) & E knn 128 auto       & 1965 (24.87\%)  & 525\\
 \holyhammer(blistr) & E knn 128 blistr   & 2556 (32.35\%) & 776\\ 
\bottomrule
\end{tabular}
 \caption{\label{tab:full_scale}
  Full-scale experiments and 
  comparison of the different strategies on a common dataset of 7902 theorems}
\end{table*}


Table~\ref{tab:full_scale} gathers the results of 4 \tactictoe strategies and 2 
\holyhammer strategies. Combining the advantages of tactic selection done by
\tactictoe with premise selection gives best results. Indeed, the combined method 
\tactictoe(SH) solves 39.42\%  on the common dataset whereas the best \holyhammer 
single-strategy only solves 32.35\% of the goals.
Surprisingly, the effect of self learning was a little negative. This may be caused
by the fact that recording both the human proof and the computer-generated script may 
cause duplication of the feature vectors which happens when the proofs are similar.
The effect of this duplication is mitigated by the orthogonalization method which 
proves 62 more theorems than the default strategy. We believe that testing even 
stronger learning schemes is one of the most crucial steps in 
improving proof automation nowadays.

Table~\ref{theories} compares the success rates of re-proving for different
\holfour theories. \tactictoe(SH) outperforms \tactictoe(NH) on every considered theory.
Even if a lot weaker due to the missing premise selection component, \tactictoe(NH)
is hugely better than \holyhammer(blistr) in 4 theories: \texttt{measure}, 
\texttt{list}, \texttt{sorting} and \texttt{finite\_map}. The main reason is that 
those theories are equipped with specialized tactics, performing complex 
transformation such as structural induction and \tactictoe(NH) can reuse them. 
Conversely, \holyhammer(blistr) is more suited to deal with dense theories such as 
\texttt{real} or \texttt{complex} where a lot of related theorems are available and 
most proofs are usually completed by rewriting tactics.

\begin{table}[b!]
\centering
\setlength{\tabcolsep}{3mm}
\begin{tabular}{@{}ccccc@{}}
\toprule
\phantom{ab} & {arith} & {real} & {compl} 
& {meas} \\
\midrule
\tactictoe(NH) & 37.3 & 19.7 & 42.6 & 19.6 \\
\tactictoe(SH) & 60.1 & 46.1 & 63.7 & 22.1 \\
\holyhammer(blistr) & 51.9 & 66.8 & 72.3 & 13.1 \\
\bottomrule
\end{tabular}

\begin{tabular}{@{}ccccc@{}}
\toprule
\phantom{abc}  & {proba} & {list} & {sort} & {f\_map} \\
\midrule
\tactictoe(NH) & 25.3 & 48.1 & 32.7 & 53.4 \\
\tactictoe(SH) & 25.3 & 51.9 & 34.7 & 55.5 \\
\holyhammer(blistr) & 25.3 & 23.3 & 16.4 & 18.1 \\
\bottomrule
\end{tabular}
\caption{\label{theories}Percentage (\%) of re-proved theorems in the theories \texttt{arithmetic}, \texttt{real}, \texttt{complex}, \texttt{measure},  
\texttt{probability}, \texttt{list}, \texttt{sorting} and \texttt{finite\_map}. }
\end{table}

\subsection{Reconstruction}
96\% of the \holyhammer(auto) (\eprover with auto strategy) proofs can be reconstructed by
\metis in 2 seconds using only the dependencies returned by the prover. In comparison, all the \tactictoe successful proofs could be reconstructed and resulted in proof scripts that were readable by \holfour and solved their goals.

Furthermore, the generated proof returned by \tactictoe is often more readable and informative than the single 
\metis call returned by \holyhammer. Since each tactic calls had a time limit of 0.02 
seconds, the reconstructed proof is guaranteed to prove the goal in a very short amount of 
time. Those considerations indicate that often \tactictoe generated scripts can
contribute to the development of formal libraries in a smoother manner.

\subsection{Time and Space Complexity}
\begin{figure}[t]
\centering
\begin{tikzpicture}[scale=0.7]
   \begin{axis}[xmin=0, xmax=5,
                ymin=1000, ymax=3200,
                width=\textwidth,
                height=0.7\textwidth,
                xtick={0,1,2,3,4,5},
                ytick={0,500,1000,1500,2000,2500,3000,3500},
                cycle list name=plotcycle,
                legend style={at={(0.97,0.03)},anchor=south east}]  
    \addplot table[x=time, y=solved] {e0_time};
    \addplot table[x=time, y=solved] {e1_time};
    \addplot table[x=time, y=solved] {ea_time};
    \legend{\tactictoe(SH),\tactictoe(NH),\holyhammer(auto)};
    \end{axis}
\end{tikzpicture}
\caption{\label{fig:time}Number of problems solved in less than $t$ seconds.} 
\end{figure}
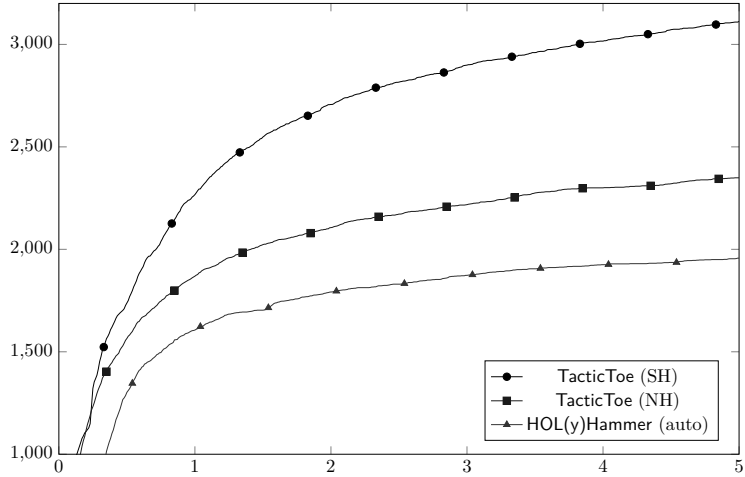

Here, we try to gain some insight by measuring different proof search variables.
We will keep track of the total number of nodes in the proof state, the time it took to get a successful proof and the size of the final proof script. In Table~\ref{tab:stats}, the number of nodes is computed over failing searches whereas the average time and the proof size is evaluated over successful searches.
We estimate the total search space explored by the number of nodes.
 It is 4 times larger in the \tactictoe(NH) version because of 
the lack of time consuming \metis calls. The average proof size (number of tactics in the final script) is around 3 even 
without explicit \metis invocations. A detailed analysis of the proofs by
their size in Fig.~\ref{fig:steps} confirms the fact that most proofs are short. One of the proofs happens to be 39 steps long, but it is not a common case. This indicates the need to focus on high-quality predictions. Since the currently recorded tactics may not cover enough space,
a way to generate new tactic calls may be necessary. 
The depiction
of the numbers of problems solved in a certain amount of time in Fig.~\ref{fig:time} shows that it is increasingly harder to solve new goals. Nevertheless, it seems that our strongest strategy \tactictoe(SH) can benefit most from an increased time limit. 

\begin{table*}[t]
\centering\ra{1.3}
\small
\begin{tabular}{llllll}
\toprule
 ID & \multicolumn{2}{c}{nodes}   & \multicolumn{2}{c}{proof size} & \multicolumn{1}{c}{time}\\ \cmidrule(lr){2-3} \cmidrule(lr){4-5} \cmidrule(lr){6-6}
     & average & max & average & max & average \\
\midrule
 \tactictoe(NH)  & 94.66 & 421 & 3.34 & 39 & 0.66 \\
 \tactictoe(SH) & 25.27 & 407 & 2.34 & 34 & 0.83 \\

\bottomrule
\end{tabular}
 \caption{\label{tab:stats}search statistics}
\end{table*}

\pgfplotscreateplotcyclelist{plotcycle}{
solid, mark repeat=50, mark phase=20, black!100, mark=*\\%
solid, mark repeat=50, mark phase=20, black!90, mark=square*\\%
solid, mark repeat=50, mark phase=20, black!80, mark=triangle*\\%
solid, mark repeat=50, mark phase=20, black!70, mark=diamond*\\%
solid, mark repeat=50, mark phase=20, black!60, mark=star\\%
solid, mark repeat=50, mark phase=20, black!50, mark=*\\%
}

\pgfplotscreateplotcyclelist{plotcycle2}{
solid, mark repeat=3, mark phase=2, black!100, mark=*\\%
solid, mark repeat=3, mark phase=2, black!90, mark=square*\\%
solid, mark repeat=3, mark phase=2, black!80, mark=triangle*\\%
solid, mark repeat=3, mark phase=2, black!70, mark=diamond*\\%
solid, mark repeat=3, mark phase=2, black!60, mark=star\\%
solid, mark repeat=3, mark phase=2, black!50, mark=*\\%
}

\begin{figure}[t]
\centering
\begin{tikzpicture}[scale=0.7]
   \begin{axis}[xmin=0, xmax=10,
                ymin=0, ymax=1500,
                width=\textwidth,
                height=0.7\textwidth,
                xtick={0,1,2,3,4,5,6,7,8,9,10},
                ytick={0,500,1000,1500,2000,2500,3000,3500},
                cycle list name=plotcycle2,
                legend style={}
                ]  
    \addplot table[x=proof_size, y=solved] {e0_step};
    \addplot table[x=proof_size, y=solved] {e1_step};
    \legend{\tactictoe(SH),\tactictoe(NH)};
    \end{axis}
\end{tikzpicture}
\caption{\label{fig:steps}Number of searches ($y$ axis) that result in a proof of size exactly $x$ ($x$ axis).} 
\end{figure}
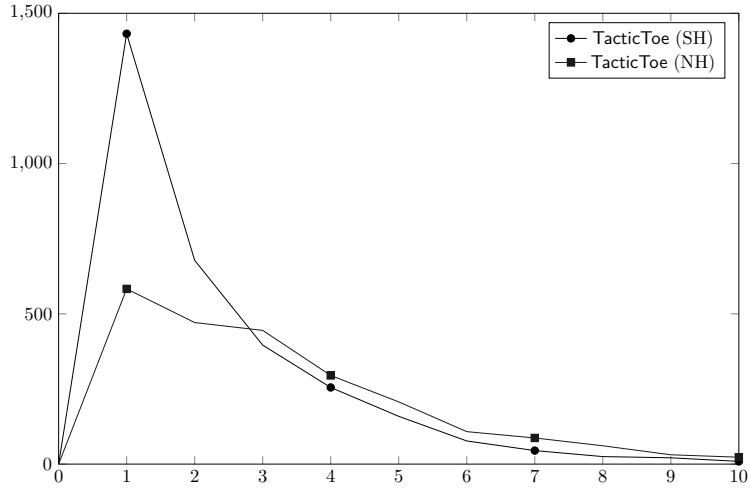  

The total time of a search is split into 5 parts:
predictions, tactic application, node creation, node selection, and node 
deletion. Usually in a failing search, the total prediction time takes less than a second, the tactic applications consume one to two seconds and the rest is used by
the node processing parts. A simple improvement would be to reduce the bookkeeping part in a future version of \tactictoe.



\subsection{Case Study}

Investigating further the different qualities of \tactictoe, we study its 
generated proof scripts on an example in \texttt{list} theory (see Example~\ref{ex:casestudy}). 
The theorem to be proven states the equivalence 
between the fact that a number $n$ is greater than the length of a list $\mathit{ls}$ with the 
fact that dropping $n$ elements from this list returns an empty list.

The human proof proceeds by induction on $n$ followed by solving both goals using rewrite steps combined with an arithmetic decision procedure. 
Both \tactictoe proofs (NH and SH) follow the general idea of reasoning by induction but solve 
the base case and the inductive case in a different way.
The base case only needs rewriting using the global simpset in the \tactictoe(NH)
proof, which is simulated by a call to \metis in the (SH) proof.
The inductive case should require an arithmetic decision procedure as hinted by the human 
proof. This is achieved by rewriting using an arithmetic simpset in the second 
proof. In the first proof however, a rewriting step and case splitting step were used
to arrive at a point where \metis calls succeed.
The tactic proof produced by \tactictoe(NH) often looks better than the one discovered by \tactictoe(SH) in that it does not involve \metis calls with a large numbers of premises. 



\begin{example}\label{ex:casestudy}(In theory \texttt{list})
\small
\begin{center}

\begin{BVerbatim}[commandchars=\\\{\}]

Goal: ``\(\forall\)ls n. (DROP n ls = []) \(\Leftrightarrow\) n \(\geq\) LENGTH ls``

Human proof: LIST_INDUCT_TAC THEN SRW_TAC [] [] THEN DECIDE_TAC

\tactictoe(NH) proof: LIST_INDUCT_TAC THENL [SRW_TAC [] [], SRW_TAC [ARITH_ss] []]

\tactictoe(SH) proof:
  LIST_INDUCT_TAC THENL 
  [ METIS_TAC [\ldots], 
    NTAC 2 GEN_TAC THEN SIMP_TAC (srw_ss ()) [] THEN 
    Cases_on `n` THENL [METIS_TAC [\ldots], METIS_TAC [\ldots]] ]
\end{BVerbatim} 
\end{center}

%
%
%
%
%
%

\end{example}

\section{Conclusion}\label{sec:concl}

We proposed a new proof assistant automation technique which combines tactic-based
proof search, with machine learning tactic prediction and a ``small hammer'' approach.
Its implementation,
\tactictoe, achieves an overall performance of 39\% theorems on the \holfour standard library surpassing \holyhammer best single-strategy and
proving 1335 additional theorems. Its effectiveness is especially visible on theories which use
inductive data structures, specialized decision procedures, and custom built simplification sets.
Thanks to the learning abilities of \tactictoe, the generated proof scripts usually
reveal the high-level structure of the proof. 
We therefore believe that predicting ITP tactics based on the current goal features is a very reasonable approach to automatically guiding proof search, and that accurate predictions can be obtained by learning from the knowledge available in today's large formal proof corpora. 

There is plenty of future work in the directions opened here.
To improve the quality of the predicted tactics, 
we would like to predict their arguments independently.
To be even more precise, the relation between the 
tactic arguments and their respective goals could be used.
Additionally, we could aim for a tighter combination with the ATP-based hammer
systems. This would perhaps make \tactictoe slower, but it might allow
finding proofs that are so far both beyond the ATPs and \tactictoe's
powers. The idea of reusing high-level blocks of reasoning and
then learning their selection could also be explored
further in various contexts. Larger frequent blocks of (instantiated) tactics
in ITPs as well as blocks of inference patterns in ATPs could be detected
automatically, their
usefulness in particular proof situations learned from the large corpora of
ITP and ATP proofs, and reused in high-level proof search.

\paragraph{Acknowledgments}\label{sect:acks}

This work has been supported by the
ERC Consolidator grant no.\ 649043 \textit{AI4REASON} and ERC starting
grant no.\ 714034 \textit{SMART}.

\bibliographystyle{plain}
\bibliography{biblio}

\appendix
\section{Recording Tactic Calls}\label{sec:appendix}

In the appendix we present the implementation details of recording tactic calls from the LCF-style
proof scripts of \holfour. We first discuss parsing the proofs and identifying individual tactic calls. We next show how tactics calls are recorded together with their corresponding goals using a modified proof script.
Finally, the recorded data is organized as feature vectors with tactics as labels and characteristics of their associated goals as features. These feature vectors constitute the training set for our selection algorithm.

\subsection{Extracting Proofs}
Our goal is to do a first-pass parsing algorithm to extract the proofs and give them to a prerecording function for further processing.
For that purpose, we create a custom theory rebuilder that parses the string representation of the theory files and modifies them. 
The proofs are extracted and stored as a list of global \sml declarations. 
The rebuilder then inserts a call to the 
prerecorder before each proof with the following arguments: 
the proof string, the string representation of each declaration occurring 
before the proof, the theorem to be proven and its name. 
After each theory file has been modified, a build of the \holfour library is 
triggered and each call of the prerecording function will perform the following steps: identifying tactics, globalizing tactics and registering tactic calls. The effects of those steps will be depicted on a running example taken from a proof in the list theory.  

\begin{example} Running example (original call)
\small
\begin{alltt}
val MAP_APPEND = store_thm ("MAP_APPEND", 
  --`!(f:'a->'b).!l1 l2. MAP f (APPEND l1 l2) = APPEND (MAP f l1) (MAP f l2)`--,
    STRIP_TAC THEN LIST_INDUCT_TAC THEN ASM_REWRITE_TAC [MAP, APPEND]);
\end{alltt}
\end{example}   

\begin{example} Running example (extracted proof)
\small
\begin{alltt}
   "STRIP_TAC THEN LIST_INDUCT_TAC THEN ASM_REWRITE_TAC [MAP, APPEND]"
\end{alltt}
\end{example}

\subsection{Identifying Tactics in a Proof}

Parsing proofs is a more complex task than extracting them due to the 
presence of infix operators with different precedences. For this reason,
in this phase we rely on the \polyml interpreter to extract tactics 
instead of building a custom parser.
In theory, recording nested tactics is possible, but we decided to restrict 
ourselves to the outermost tactics, excluding those constructed by a tactical
(see list in Example~\ref{ex:tactical}).
The choice of the recording level was made to reduce the complexity of 
predicting the right tactic and minimizing the number of 
branches in 
the proof search. In particular, we do not consider \texttt{REPEAT} to be a 
tactical and record  \texttt{REPEAT X} instead of repeated calls to \texttt{X}.
  
\begin{example}\label{ex:tactical}
\small
\begin{alltt}
THEN ORELSE THEN1 THENL REVERSE VALID by suffices\_by
\end{alltt}
\end{example} 
  
\begin{example} Running example (identified tactics)
\small
\begin{alltt}
"STRIP_TAC" "LIST_INDUCT_TAC" "ASM_REWRITE_TAC [MAP, APPEND]"
\end{alltt}
\end{example}   

\subsection{Globalizing Tactics}

The globalization process attempts to modify a tactic string so that it is 
understood by the compiler in the same way anywhere during the build of 
\holfour. In that manner, the \tactictoe proof search will be able to reuse 
previously called tactics in future searches. A first reason why a tactic may 
become inaccessible is that the module where it was declared is not open in 
the current theory. Therefore, during the prerecording we call the \polyml 
compiler again to obtain the module name (signature in \sml) of the tactic 
tokens.
This prefixing also avoids conflicts, where different tactics with the name 
appears in different module. There are however some special cases where the 
module of a token is not declared in a previous module.
If the token is a string, already prefixed, or a 
\sml reserved token then we do not need to perform any modifications. If a 
value is declared in the current theory (which is also a module), we replace 
the current value (or function) by its previous declaration in the file. This
is done recursively to globalize the values. Theorems are treated in a special 
manner. Thanks to the \holyhammer tagging system~\cite{holyhammer}, they can be in most cases 
fetched from the \holfour 
database. Terms are reprinted with their types to avoid misinterpretation 
of overloaded constants.
 
 
Since certain values in \holfour are stateful (mostly references), we cannot guarantee 
that the application of a tactic will have exactly the same effect in a 
different context. This is not a common issue, as a fully functional style is
preferred, however there is one important stateful structure that we need to
address: the simplification set is stored globally and the simplification procedures
rely on the latest version available at the moment of the proof.
 
\begin{example} Running example (globalized tactics)

The tactic \texttt{LIST\_INDUCT\_TAC} is not defined in the signature of the $list$ theory. That is why, to be accessible in other theories its definition appears in its globalization.

\begin{alltt}
"Tactic.STRIP_TAC"
"let val LIST_INDUCT_TAC = Prim_rec.INDUCT_THEN 
( DB.fetch \bq{list} \bq{list_INDUCT} ) Tactic.ASSUME_TAC in LIST_INDUCT_TAC end"
"Rewrite.ASM_REWRITE_TAC 
[( DB.fetch \bq{list} \bq{MAP} ) , ( DB.fetch \bq{list} \bq{APPEND} ) ]"
\end{alltt}
\end{example}   
 
\subsection{Registering Tactic Calls}
To judge the effectiveness of a tactic on proposed goals, we record how it performed previously in similar situations. For that, we modify the proofs to record and associate the globalized tactic with the goal which the original 
tactic received. Each original tactic is automatically modified to
perform this recording as a side effect. The code of the record function \texttt{R} is defined below in Example~\ref{ex:record}. The first line 
checks if the globalized tactic \texttt{gtac} 
produces the same open goals as the original tactic. 
In the second line we save the globalized tactic and the features of the goal 
to a file. Storing features instead of goals was preferred in order to 
avoid unnecessary recomputation. It is also more convenient since features can 
be stored as a list of strings. In the running example only constant features are presented (the complete set of extracted features was discussed in Section~\ref{sec:features}). Finally, the original tactic is called to continue the proof.

\begin{example}\label{ex:record} Pseudo-code of the recording function
\begin{alltt}
fun R (tac,gtac) goal = 
  (test_same_effect gtac tac goal; save (gtac, features_of goal); tac goal)
\end{alltt}
\end{example}

\begin{example} Running example (recording proof string)
R is the recording function
\small
\begin{alltt}
( ( R ( STRIP_TAC , "Tactic.STRIP_TAC" ) ) ) THEN 
( ( R ( LIST_INDUCT_TAC , "( let val LIST_INDUCT_TAC = Prim_rec.INDUCT_THEN 
( DB.fetch \bq{list} \bq{INDUCT}) Tactic.ASSUME_TAC in LIST_INDUCT_TAC end )" ) ) ) THEN 
( R ( ASM_REWRITE_TAC [ MAP , APPEND ] , "Rewrite.ASM_REWRITE_TAC 
[( DB.fetch \bq{list} \bq{MAP} ) , ( DB.fetch \bq{list} \bq{APPEND} ) ]" ) )
\end{alltt}
\end{example}   

The application of the recording function \texttt{R} and its subcalls will 
only take place during a second \holfour build where the 
proofs have been replaced by their recording variants. This replacement will be performed by a 
modified version of the rebuilder that extracted the proofs. It will also 
create a call to our search algorithm before 
the recording proof and a call to a post-recorder after it. The post-recorder 
will create feature vectors consisting of the name of the 
current theorem, its features and every globalized tactic in the proof. This second set
is used to preselect the tactics before trying to 
re-prove a theorem (see Section~\ref{sec:pretactic}). 
\end{document}